\documentclass[10pt,letterpaper]{article}
\usepackage{envrethink-arxiv}
\usepackage[T1]{fontenc}
\usepackage{lmodern}
\usepackage[numbers,square,comma,sort&compress]{natbib}
\usepackage{amsmath,amssymb}
\usepackage{hyperref}
\usepackage{url}
\usepackage{hyphenat}
\usepackage{xspace}
\usepackage{enumitem}
\usepackage{tabularx}
\usepackage{pdflscape}
\usepackage{longtable}
\usepackage{flafter}
\usepackage{tikz}
\usepackage{graphicx}
\usepackage{booktabs}
\usepackage{array}
\usepackage{float}
\usepackage{colortbl}
\usepackage{xcolor}
\usepackage{siunitx}   
\usepackage{multirow}
\usepackage[table]{xcolor}
\usepackage{fontawesome}
\definecolor{headbg}{gray}{0.90}
\definecolor{rowA}{gray}{0.965}
\definecolor{rowB}{gray}{0.985}
\definecolor{tblgroup}{RGB}{230,240,252}
\definecolor{tblgain}{RGB}{0,150,70}
\definecolor{tblloss}{RGB}{215,40,40}

\usepackage{placeins}

\usetikzlibrary{arrows.meta,positioning,fit,shapes.geometric,calc}
\title{AmbiHTMLBench: A Benchmark for HTML-based Visually Rich Document parsing}
\newcommand{\shortpapertitle}{AmbiHTMLBench}
\newcommand{\preprintlogos}{%
  \raisebox{-0.5\height}{\includegraphics[width=0.32\linewidth]{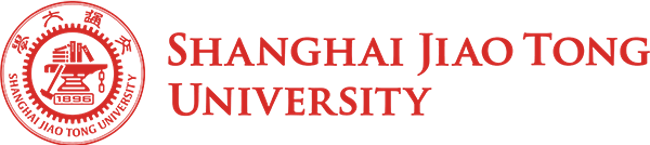}}\hfill
  \raisebox{-0.5\height}{\includegraphics[width=0.32\linewidth]{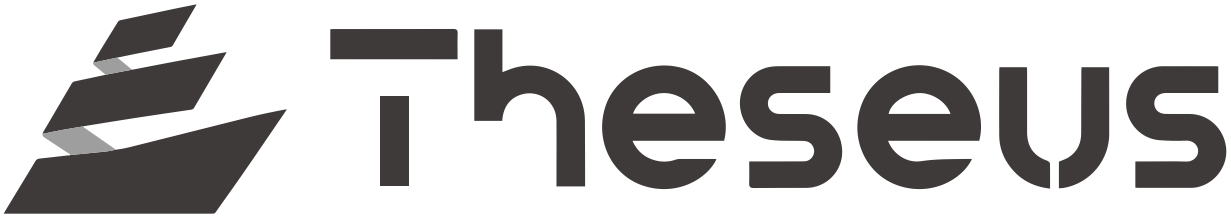}}%
}
\newcommand{\paperauthors}{%
  \mbox{Boyu Niu\textsuperscript{1,2,3}}\quad
  \mbox{Yingjie Xiong\textsuperscript{1,2}}\quad
  \mbox{Shurui Li\textsuperscript{1,2}}\quad
  \mbox{Linhan Pang\textsuperscript{1,2}}\quad
  \mbox{Yuntong Zhang\textsuperscript{1,2}}\\[0.3em]
  \mbox{Han Wang\textsuperscript{3}}\quad
  \mbox{Hao Wu\textsuperscript{3}}\quad
  \mbox{Kai Zuo\textsuperscript{3}}\quad
  \mbox{Xu Tang\textsuperscript{3}}\quad
  \mbox{Xuanhe Zhou\textsuperscript{1,2}\thanks{Corresponding author.}}\quad
  \mbox{Fan Wu\textsuperscript{1,2}}
}
\author{%
  \paperauthors\\[0.5em]
  {\small\textsuperscript{1}Shanghai Jiao Tong University\quad
  \textsuperscript{2}Theseus Labs\quad
  \textsuperscript{3}Xiaohongshu}\\[0.2em]
  \href{https://ambihtmlbench.tech}{\faGlobe\ Homepage}\hspace{1.5em}
  \href{https://github.com/theseus-labs-rsi/AmbiHTMLBench}{\faGithub\ GitHub}\hspace{1.5em}
  \href{https://huggingface.co/datasets/Theseus-Lab/AmbiHTMLBench}{\faCode\ Hugging Face}
  \vspace{-1.5em}
}
\date{}

\begin{document}
\maketitle
\begin{abstract}
Traditional document parsing systems use Markdown as their output representation, which is effective for linear text but does not fully preserve fine-grained spatial structure, table geometry, visual hierarchy, or rendering semantics. We show that HTML is advantageous for heterogeneous and layout-sensitive documents whose representations must remain editable, inspectable, and reusable, and introduce AmbiHTMLBench to systematically study this setting. In this benchmark, given a real-world document, a model must recover executable HTML that preserves its content, organization, and visual appearance. The benchmark contains 1,115 diverse pages with structured annotations and evaluates predictions in two different ways: The objective protocol aligns ground-truth and predicted rendered blocks and measures table fidelity, page-level text error, and a two-dimensional layout consistency; a complementary reference-anchored protocol measures perceived fidelity by aligning LLM judges with human preferences. Across ten primary multimodal systems, the best objective score is 0.8092, while the subjective winner is different, demonstrating that automatic decomposition and human-perceived fidelity capture complementary properties. We further fine-tune a small-parameter expert model and find that task-specific adaptation yields good performance on document-to-HTML reconstruction task. These results position AmbiHTMLBench as a diagnostic testbed for building smarter and more logical document parsing systems.

\end{abstract}

\begin{figure}[htbp]
\centering
\includegraphics[width=0.75\linewidth]{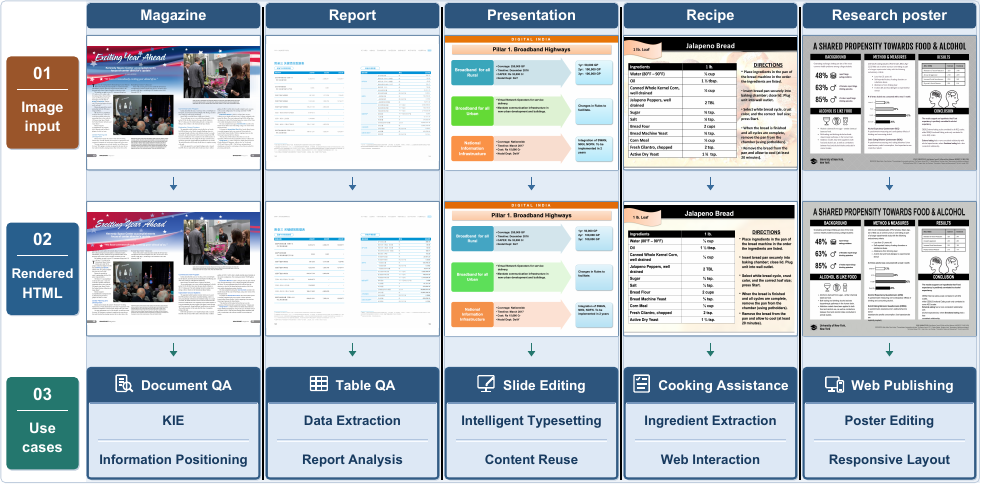}
\caption{\textbf{From document images to reusable HTML.} Complete pages from realistic genres (top) are reconstructed as
HTML and rendered in browser (middle), more cases are shown in appendix \ref{app:case study}. The bottom row lists potential uses of the generated content.}
\label{fig:overview}
\end{figure}

\section{Introduction}
\begin{figure}[htbp]
\centering
\includegraphics[width=0.86\linewidth]{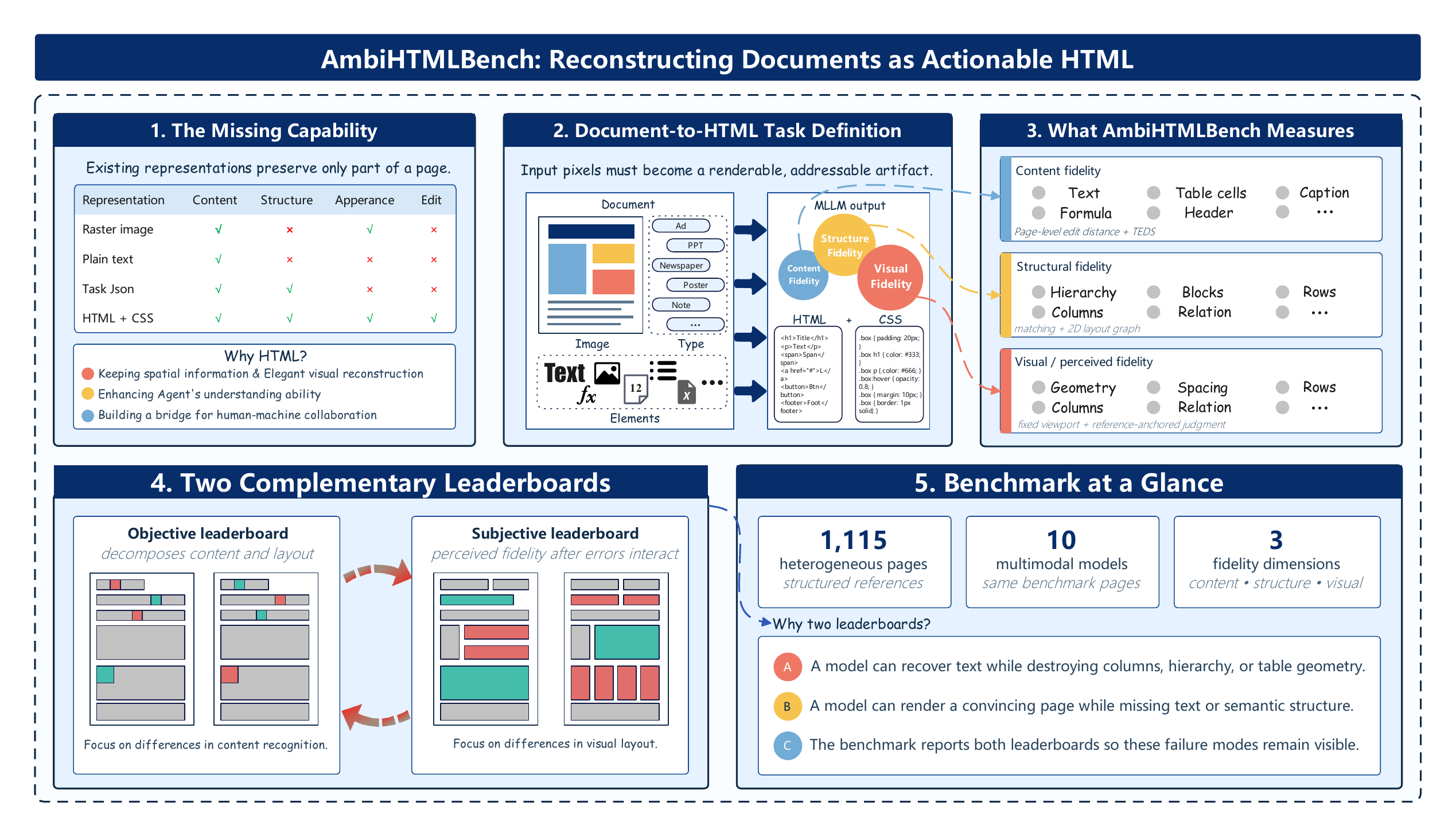}
\caption{AmbiHTMLBench focuses on three level of fidelity in HTML based document parsing, with both objective and reference-anchored subjective protocols.}
\label{fig:overview}
\end{figure}

\paragraph{Motivation.} Documents are not merely sequences of words: headings, tables, figures, typography, and spatial arrangement jointly determine how information is interpreted. Recent multimodal systems can extract and reason over document images \citep{xu-etal-2021-layoutlmv2,wang-etal-2024-docllm,hu-etal-2024-mplug}, but document agents require a stronger capability. They must be able to store evidence, locate the source of an answer, modify a document, and reuse the same representation across retrieval, reasoning, and interaction tasks \citep{han2025mdocagent}. Understanding a page once is therefore insufficient; agents need a persistent and actionable representation of the page itself.

Existing representations expose only part of this information. Plain text or Markdown supports efficient reading but discards hierarchy, spatial relations, and visual emphasis. JSON-like schemas can encode selected semantics, yet are usually designed for one downstream task and rarely preserve presentation. Raster images retain appearance but provide neither editable semantic units nor stable machine-addressable references. These limitations force downstream agents to repeatedly infer structure from pixels, creating a gap between document perception and document interaction.

HTML provides a natural representation for closing this gap. Its Document Object Model (DOM) expresses hierarchical and addressable relationships among headings, paragraphs, tables, figures, and captions, allowing humans and agents to locate and manipulate individual elements \citep{li-etal-2022-markuplm,deng2022domlm}. At the same time, HTML and CSS provide an executable rendering mechanism for alignment, spacing, typography, positioning, and table geometry. Semantic elements and their visual realization remain connected rather than being separated into an image and an unrelated text transcript. A faithful document-to-HTML system should thus produce an artifact that is simultaneously readable, editable, inspectable, and renderable.

We also obtain direct evidence for this interface property in two small pilots. On six field groups from a UKSwap form, HTML/CSS yields 62.03\% pair-joint accuracy for field-to-slot binding, compared with 44.44\% for geometry-matched non-HTML records. On PubTabNet tables, HTML-based evidence summaries make row, column, and value nodes explicitly addressable, enabling more reliable correction of an erroneous row assignment than linearized Markdown. These pilots are not part of the benchmark leaderboard; they motivate evaluating whether models can reconstruct the representation that agents can subsequently inspect and act upon.

\begin{figure}[htbp]
\centering
\includegraphics[width=0.84\linewidth]{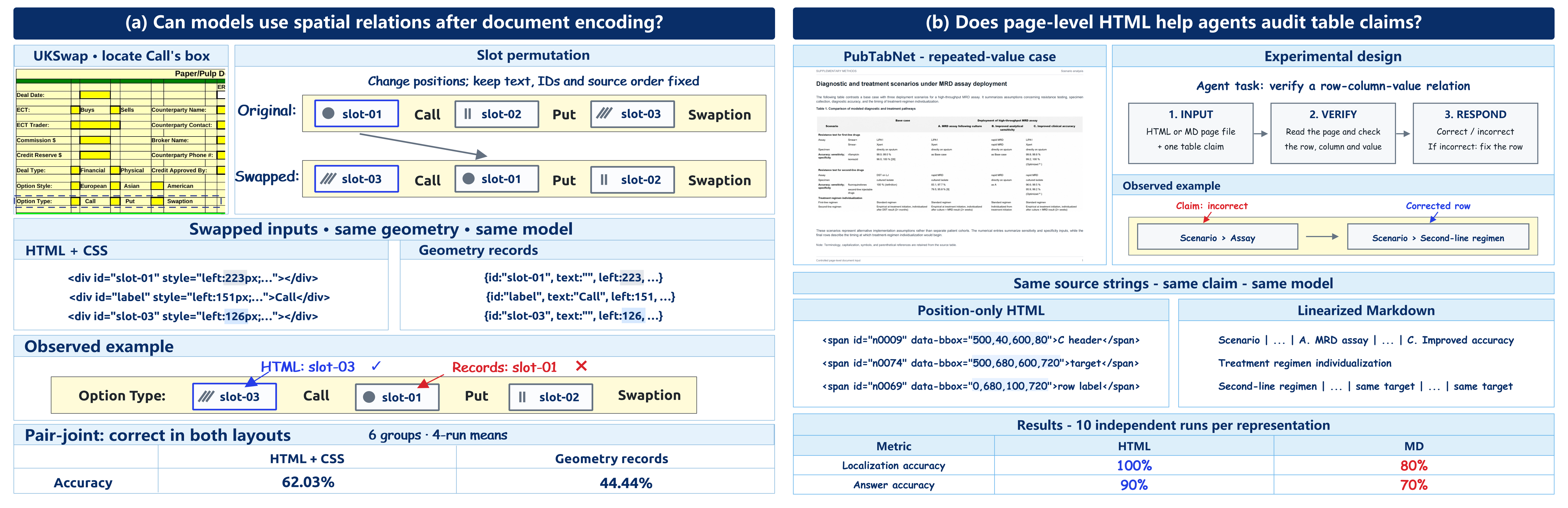}
\caption{Pilot evidence that HTML representations support spatial binding and table verification.}
\label{fig:html-preexperiments}
\end{figure}

\paragraph{The missing capability.} Despite progress in OCR and document parsing, existing benchmarks largely measure isolated components such as text, tables, formulas, and reading order \citep{Ouyang_2025_CVPR,zhang2026parsebench,li2026puredocbench}. Screenshot-to-code benchmarks study visual reconstruction, but primarily target webpages with relatively regular DOM-oriented layouts. Neither setting directly tests whether a multimodal model can recover a heterogeneous real-world document as complete, executable HTML. This task is difficult because fidelity has several interacting dimensions: a prediction may contain the correct words but destroy columns or table geometry; it may preserve layout while hallucinating or omitting content; or it may encode a valid DOM whose CSS fails to reproduce the visual hierarchy that guides human reading.

These failure modes also motivate complementary evaluation. Content fidelity should measure whether text, numbers, formulas, and table cells are preserved. Structural fidelity should measure whether blocks and their relationships are organized as in the source. Visual or perceived fidelity should account for the fact that humans may tolerate harmless line-break differences but strongly penalize lost emphasis, displaced media, or broken table structure. A single metric cannot reliably expose all of these cases, so a useful benchmark should report decomposed automatic signals together with a reference-anchored perceptual measure.

\paragraph{Our benchmark.} We present \textbf{AmbiHTMLBench}, a benchmark of 1,115 diverse pages with expert annotations. We render each generated HTML page and align its semantic blocks with the ground truth to measure page-level text error, table fidelity, and two-dimensional layout consistency. We additionally use human-calibrated VLM judgments to estimate perceived fidelity under a separate subjective protocol. This design lets us ask not only whether a model obtains a high score, but also which reconstruction capability is responsible for success or failure.

Our contributions are threefold: (1) a diverse document-to-HTML benchmark covering heterogeneous page types, content, and layouts; (2) a reproducible multi-dimensional evaluation protocol that separates content, structure, rendering, and perceived fidelity; and (3) an empirical study of ten multimodal systems and task-specific fine-tuning. The objective and subjective leaderboards select different winners, and adaptation improves smaller models, showing that AmbiHTMLBench is useful both as a diagnostic evaluation and as a training target for agent-ready document representations.

\section{Related Work}

Document understanding has progressed from OCR and layout analysis to multimodal models that jointly encode text, appearance, and spatial position, including LayoutLM variants, DocFormer, LiLT, Donut, DocLLM, and mPLUG-DocOwl \citep{Xu_2020,xu2022layoutlmv2multimodalpretrainingvisuallyrich,huang2022layoutlmv3pretrainingdocumentai,appalaraju2021docformerendtoendtransformerdocument,wang2022liltsimpleeffectivelanguageindependent,kim2022ocrfreedocumentunderstandingtransformer,wang2023docllmlayoutawaregenerativelanguage,ye2023mplugdocowlmodularizedmultimodallarge}. Recent agent frameworks add retrieval and tool use, but generally reason over extracted content rather than reconstructing complete documents \citep{han2025mdocagentmultimodalmultiagentframework,yang2025docagentmultiagentautomatedcode}.

Visual generation work such as Pix2Struct, AnyDoc, WebSight, and Design2Code maps screenshots to executable markup, primarily for webpages \citep{pmlr-v202-lee23g,lin2026anydoc,laurençon2024unlockingconversionwebscreenshots,si-etal-2025-design2code}. Conversely, OmniDocBench, PureDocBench, and ParseBench assess document parsing and component-level fidelity \citep{Ouyang_2025_CVPR,li2026puredocbench,zhang2026parsebench}. AmbiHTMLBench bridges these directions by evaluating full document-to-HTML reconstruction, with content, layout, and perceived fidelity measured separately.

\section{Benchmark Construction}

\begin{figure}[htbp]
    \centering
    \includegraphics[width=0.88\linewidth]{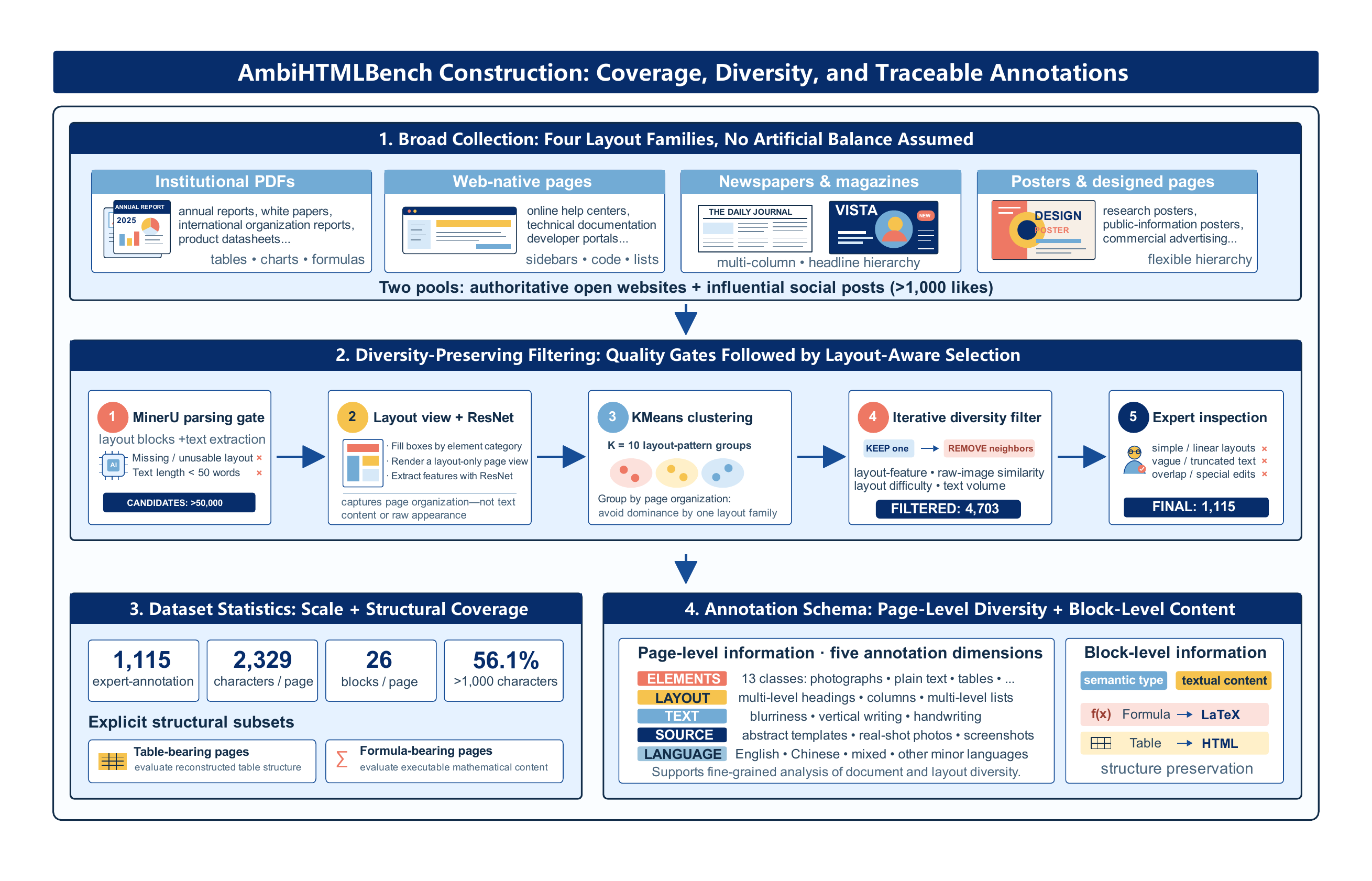}
    \caption{The overall AmbiHTMLBench Construction process.}
    \label{fig:benchmark construction}
\end{figure}

We collected heterogeneous document images and applied multi-stage quality control to preserve diversity in document type, style, and layout while reducing near-duplicates.

  \subsection{Data Collection}

Sources span four categories: institutional PDFs (e.g., reports and datasheets), web-native documentation, newspapers and magazines, and posters or other strongly designed pages. Together they cover conventional and nonstandard layouts, dense text, tables, figures, and mixed media.

\subsection{Filtering and Quality Control}
After data collection, there are above 50,000 pages remaining. Before producing the final benchmark subset, we applied several quality control steps. First, we use MinerU\citep{niu2025mineru25decoupledvisionlanguagemodel} to conduct basic document parsing, and (1) Remove samples whose layout parsing result is missing or unusable. (2) Remove samples with less than 50 words. This step ensures that every retained sample has both a clear layout structure and abundant text content. 

Next, based on MinerU layout parsing result, we provide a layout view of the documents by fill the bounding boxes with different colors according to their categories, and then use a layout-aware clustering to reduce redundancy. All candidate samples are clustered into ten groups using KMeans over the features obtained after feature extraction of this layout view using ResNet-50. This process groups documents primarily by page organization rather than by raw appearance or textual content.

Within each cluster, we applied an iterative diversity-preserving filtering procedure. At each iteration, we retained one candidate and removed nearby pages according to layout-feature similarity, visual similarity, estimated layout difficulty, and text volume. This procedure was repeated until the retained set reached the desired diversity level.

This procedure balances diversity and quality by limiting dominant layout families, removing near-duplicates, and retaining informative pages across difficulty levels.

After filtering 50,000+ candidates to 4,703 pages, expert review removed overly simple, truncated, overlapping, or specially edited samples, yielding 1,115 benchmark pages.
\subsection{Dataset Statistics}

Experts annotated all 1,115 pages at page and block levels. Page-level labels cover 13 element types, layout (headings, columns, lists), text appearance (blur, vertical writing, handwriting), source type, and language. Block annotations record type and text; formulas use LaTeX and tables use HTML. Pages average 2,329 characters and 26 blocks, and 56.1\% exceed 1,000 characters.

\section{Evaluation Protocol}

\subsection{Objective Evaluation}
\label{sec:objective-evaluation}

We render each prediction in headless Chromium at a fixed viewport and extract visible text, lists, tables, figures, formulas, and normalized bounding boxes. Ground-truth annotations are converted to the same block representation, enabling source-independent comparison of content and geometry.

Each object is normalized as
\[
o=(k,t,b),
\]
where \(k\) is its category, \(t\) its textual or structural content, and \(b=(x,y,w,h)\) its page-normalized bounding box.

\subsubsection{Block Matching Algorithm}

Tables are matched first as table objects. Regular text, formulas, and cells from unmatched tables then share a text pool, giving text credit to linearized tables while reserving structural credit for preserved table objects.

Table candidates combine content, structure, text, and geometry scores and are assigned one-to-one by thresholded Hungarian matching.

Because HTML may split or merge annotated text blocks, partial text similarity defines a bipartite candidate graph; connected components support one-to-many alignments, with residuals retained as insertions or deletions. The resulting shared alignment feeds both page-level text scoring and layout analysis. Details are in Appendix~\ref{app:objective-details}.

\subsubsection{Content consistency}

For matched tables, we report normalized row/cell edit scores: TEDS includes cell text, while TEDS-S measures structure only (an approximation, not full DOM tree edit):
\[
S_{\mathrm{table}}=1-\frac{d_{\mathrm{row/cell}}}
{\max(|T_g|,|T_p|,1)}.
\]
Scores are reported on the 263 pages containing tables.

Page-level text fidelity concatenates aligned groups, including one-sided residuals, and computes normalized character edit distance:
\[
D_{\mathrm{text}} =
\frac{\operatorname{Lev}(T_g,T_p)}
{\max(|T_g|,|T_p|,1)} .
\]

\subsubsection{Layout consistency}
	
Layout uses matched tables and text groups as nodes, compares normalized center and size, and evaluates corresponding edges in a ground-truth 3-nearest-neighbor graph. Empty-sided or unsupported text matches do not receive layout credit.

For an edge, similarity combines direction, relative displacement vector, and length:
\[
s_{\text{edge}} =
0.4s_{\text{dir}}+0.3s_{\text{len}}+0.3s_{\text{vec}}.
\]
The page-level layout score is
\[
S_{\text{layout}} =
0.55\,\overline{s}_{\text{edge}}
+0.30\,\overline{s}_{\text{node}}
+0.15F_1^{\text{coverage}}.
\]
The score retains two-dimensional direction and displacement; coverage penalizes omitted reference content.

\subsection{Subjective Evaluation}
\label{sec:subjective-evaluation}

Objective metrics cannot fully capture how content and layout errors interact in perception, so we add reference-anchored pairwise judgments.

\begin{figure}[htbp]
\centering
\includegraphics[width=0.82\linewidth]{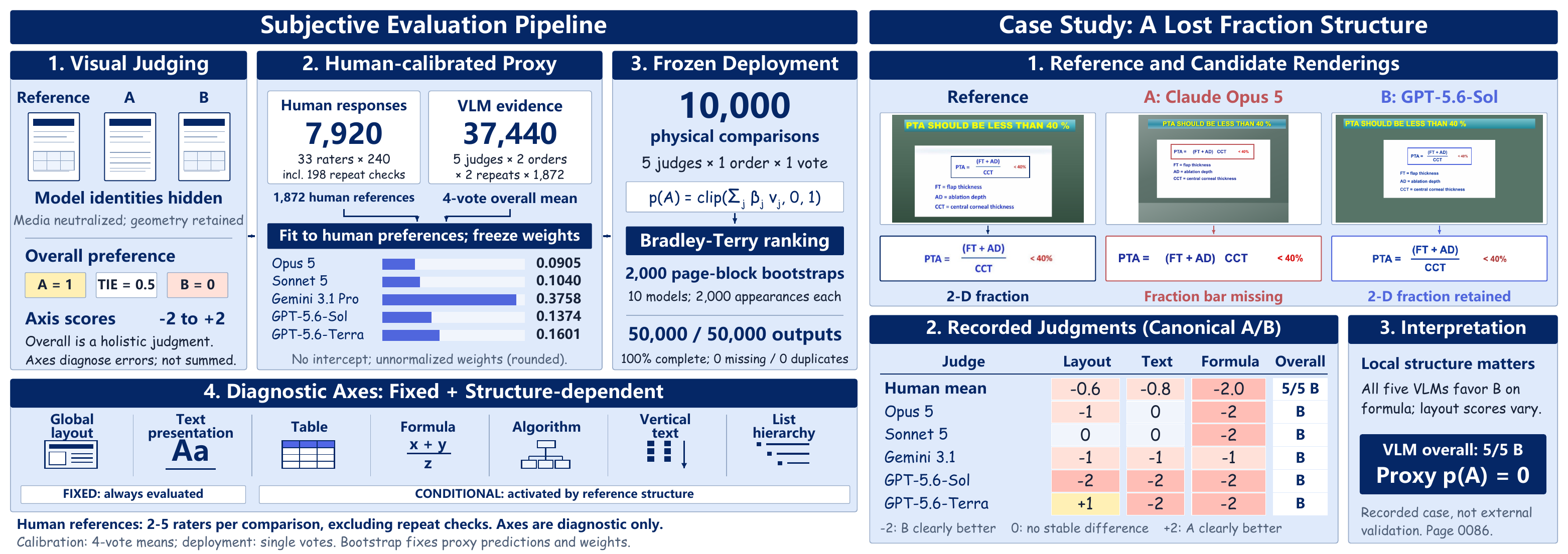}
\caption{Human calibration and deployment of the frozen preference proxy.}
\label{fig:subjective-evaluation-pipeline}
\end{figure}

Each trial presents a reference page and two reconstructions and asks which is more faithful. Figure~\ref{fig:subjective-evaluation-pipeline} summarizes the protocol.

\paragraph{Judgement protocol and human reference.}
Judges choose A, B, or tie and provide confidence and rationale. Photograph and chart interiors are neutralized while presence, geometry, and placement remain visible.

Diagnostic axes cover global layout, text presentation, and element-specific properties (tables, formulas, algorithms, vertical text, and list hierarchy); scores range from $-2$ to $2$ and do not affect the overall ranking.

Thirty-three participants completed 7,920 pairwise judgments (7,722 main annotations and 198 hidden repeats; repeats are excluded from calibration; see Appendix~\ref{app:human-study}). We encode A, tie, and B as 1, 0.5, and 0; the mean valid vote is
\[
h_i=\frac{n_{i,A}+0.5n_{i,T}}{n_{i,A}+n_{i,T}+n_{i,B}}.
\]
$n_{i,A}$, $n_{i,T}$, and $n_{i,B}$ are valid vote counts.

\paragraph{Frozen proxy and deployment.}
Five VLM judges provide canonicalized votes $x_{i,j}$. We fit and freeze a linear proxy on human-calibrated comparisons:
\begin{align}
\hat p_i(A) = \operatorname{clip}_{[0,1]}\left(\sum_{j=1}^{5}\beta_j x_{i,j}\right),
\label{eq:ols5-proxy}
\end{align}
Coefficients are estimated by ordinary least squares on human labels only. A centered Bradley--Terry model uses $\hat p_i(A)$ as a fractional outcome; page-block bootstrap estimates ranking uncertainty. The schedule includes 1,872 human-labelled and 8,128 proxy-scored comparisons. Prompts, audits, coefficients, and implementation details appear in Appendix~\ref{app:subjective}.

\section{Experiments}

We organize the experiments around two complementary leaderboards. The objective leaderboard decomposes reconstruction quality into content and layout fidelity, whereas the subjective leaderboard captures perceived fidelity after these errors interact.

\subsection{Experimental Setup}

We evaluate fifteen multimodal models on the same 1,115-page benchmark set. Each model receives the page image under its standard image-to-HTML prompting interface, and only successfully rendered outputs are scored. Objective scores are computed from the fixed protocol in Section~\ref{sec:objective-evaluation}; the subjective leaderboard uses the frozen preference proxy in Section~\ref{sec:subjective-evaluation}. We merge the final reruns of previously identified low-scoring pages before producing the reported leaderboard. Rendering, matching thresholds, and aggregation are fixed before comparing models; implementation details are provided in Appendix~\ref{app:objective-details}.

\subsection{Objective Leaderboard}

\begin{table}[t]
  \caption{Objective leaderboard on 1{,}115 pages for fifteen primary systems.
  Text ED is lower-is-better; all other columns are higher-is-better.
  The weighted score uses table weight 0.3 and text and layout weights 1.0.
  TEDS and TEDS-S are averaged over the 263 table-bearing pages.}
  \label{tab:objective-leaderboard}
  \centering
  \small
  \setlength{\tabcolsep}{5pt}
  \renewcommand{\arraystretch}{1.12}

  \begin{tabularx}{\linewidth}{@{} c >{\raggedright\arraybackslash}X
      S[table-format=1.4] S[table-format=1.4] S[table-format=1.4]
      S[table-format=1.4] S[table-format=1.4] @{}}
    \toprule
    \rowcolor{headbg}
    \textbf{Rank} & \textbf{Model}
      & \textbf{Text ED} $\downarrow$
      & \textbf{TEDS} $\uparrow$
      & \textbf{TEDS-S} $\uparrow$
      & \textbf{Layout} $\uparrow$
      & \textbf{Obj.\ score} $\uparrow$ \\
    \midrule
    \rowcolor{rowA} 1  & Gemini 3.1 Pro Preview & \bfseries 0.1754 & \bfseries 0.6518 & \bfseries 0.6650 & 0.8410 & \bfseries 0.8092 \\
    \rowcolor{rowB} 2  & Claude Opus 5.5        & 0.1994 & 0.6433 & 0.6553 & 0.8574 & 0.8048 \\
    \rowcolor{rowA} 3  & Gemini 3.8 Flash       & 0.2070 & 0.6415 & 0.6530 & 0.8288 & 0.7888 \\
    \rowcolor{rowB} 4  & Kimi-K3                & 0.2399 & 0.6253 & 0.6389 & 0.8307 & 0.7732 \\
    \rowcolor{rowA} 5  & Claude Opus 5          & 0.2537 & 0.6290 & 0.6465 & 0.8387 & 0.7712 \\
    \rowcolor{rowB} 6  & Qwen 3.8 max           & 0.2470 & 0.6095 & 0.6235 & 0.8198 & 0.7633 \\
    \rowcolor{rowA} 7  & GPT 6.1 sol            & 0.3082 & 0.6277 & 0.6390 & \bfseries 0.8654 & 0.7589 \\
    \rowcolor{rowB} 8  & GPT 6 astra            & 0.3133 & 0.6188 & 0.6324 & \bfseries 0.8654 & 0.7555 \\
    \rowcolor{rowA} 9  & Qwen3.7-Plus           & 0.2659 & 0.6113 & 0.6250 & 0.8088 & 0.7505 \\
    \rowcolor{rowB} 10 & GPT 5.6 Sol            & 0.3055 & 0.5967 & 0.6103 & 0.8425 & 0.7461 \\
    \rowcolor{rowA} 11 & GPT 5.6 Terra          & 0.2998 & 0.5953 & 0.6131 & 0.8335 & 0.7445 \\
    \rowcolor{rowB} 12 & Claude Sonnet 5        & 0.2935 & 0.6042 & 0.6218 & 0.8218 & 0.7433 \\
    \rowcolor{rowA} 13 & ERNIE 5.0              & 0.2871 & 0.5982 & 0.6137 & 0.8120 & 0.7410 \\
    \rowcolor{rowB} 14 & GLM 5V Turbo           & 0.3081 & 0.6121 & 0.6316 & 0.8103 & 0.7330 \\
    \rowcolor{rowA} 15 & Grok 4.5               & 0.6754 & 0.2874 & 0.3024 & 0.3950 & 0.3504 \\
    \bottomrule
  \end{tabularx}
\end{table}

Gemini 3.1 Pro Preview leads the objective ranking, primarily because it has the lowest page-level text error and the strongest table score while maintaining a competitive layout score. Kimi-K3 and Claude Opus 5 form the next tier: Kimi-K3 is slightly stronger on text, whereas Claude Opus 5 is slightly stronger on layout and table structure. Qwen3.7-Plus ranks fourth by balancing moderate text error with stable table and layout scores. GPT-5.6-Sol obtains the highest layout score, but its larger text error and weaker table score place it below the content-leading systems. Grok-4.5 is separated from the remaining models on every axis, especially text, table fidelity, and layout. These results motivate reporting the decomposition rather than interpreting a single score as a complete description of reconstruction quality.

\subsection{Task-Specific Fine-tuning}
\label{sec:finetuning}

To test whether AmbiHTMLBench can also serve as a training signal, we fine-tune Qwen3.5-9B and Qwen3.5-4B with transcription-style distillation. We use 1,230 accepted examples from 1,475 images in the PureDocBench \texttt{real\_degraded} track, with no overlap with the benchmark pages. Gemini 3.1 Pro Preview produces HTML using the same transcription instruction as inference; outputs are retained using a block-recall filter (83.4\% acceptance), with best-of-four resampling for initially low-scoring pages. This construction favors faithful transcription of the observed page rather than authored or normalized HTML.

Both model sizes use the same one-epoch LoRA recipe: rank and alpha 16, all linear language-model layers trainable, frozen vision and alignment modules, learning rate $10^{-5}$ with cosine decay, and a maximum sequence length of 21,504. Checkpoints are first screened with inexpensive coverage and output-discipline diagnostics and then selected by the full objective evaluator. Table~\ref{tab:finetuning-results} reports the best checkpoints on all 1,115 pages.

\begin{table}[t]
  \caption{Task-specific fine-tuning results on AmbiHTMLBench.
  All reported metrics are higher-is-better; ``Text'' is the normalized
  text score and the weighted score is the primary objective metric.}
  \label{tab:finetuning-results}
  \centering
  \small
  \setlength{\tabcolsep}{6pt}
  \renewcommand{\arraystretch}{1.12}

  \begin{tabular}{@{} l l
      S[table-format=1.4] S[table-format=1.4]
      S[table-format=1.4] S[table-format=1.4] @{}}
    \toprule
    \rowcolor{headbg}
    \textbf{Model} & \textbf{Method}
      & \textbf{Weighted} $\uparrow$
      & \textbf{Text} $\uparrow$
      & \textbf{TEDS} $\uparrow$
      & \textbf{Layout} $\uparrow$ \\
    \midrule
    \rowcolor{rowA}
    Qwen3.5-9B & Base
      & 0.6623 & 0.6321 & 0.5440 & 0.7281 \\
    \rowcolor{rowB}
    Qwen3.5-9B & Distill SFT
      & \bfseries 0.7064 & \bfseries 0.6761 & \bfseries 0.5827 & \bfseries 0.7737 \\
    \midrule
    \rowcolor{rowA}
    Qwen3.5-4B & Base
      & 0.6786 & 0.6323 & 0.5382 & 0.7670 \\
    \rowcolor{rowB}
    Qwen3.5-4B & Distill SFT
      & \bfseries 0.6934 & \bfseries 0.6556 & \bfseries 0.5722 & \bfseries 0.7675 \\
    \bottomrule
  \end{tabular}
\end{table}

Distillation improves the 9B and 4B bases by 4.41 and 1.48 weighted points, respectively, with gains on every 9B axis and on text and table fidelity for 4B. The smaller adapted model also exceeds the unadapted 9B base (0.6934 versus 0.6623). A matched-budget 27B ablation isolates the target style: transcription-style targets reach 0.7373 from a 0.7320 base, whereas cleaned, normalized, and original authored-HTML targets reach 0.7102, 0.7141, and 0.6850. Thus, the alignment between the supervision format and the reconstruction task, rather than surface cleanliness alone, determines the direction of the SFT gain. A subsequent DAPO-style RL stage adds only 0.16 and 0.27 points over the best 9B and 4B SFT checkpoints, respectively, which is within the observed evaluation variability; we therefore use SFT as the primary adapted-model result.

\subsection{Reliability of Subjective Evaluation}

\paragraph{Human reliability.}
Valid A/B/TIE preferences cover 98.85\% of the 7,722 main annotations and provide a reference for all 1,872 comparisons. Exact inter-rater agreement is 62.30\% (95\% CI: 60.79--63.82), while within-rater agreement on 198 hidden-repeat pairs is 73.23\%. Despite these disagreements, removing any one of the 33 participants leaves the ten-model human ranking unchanged (Spearman $=1.0$; maximum rank movement $=0$). This supports stability to participant removal, not unanimity on individual comparisons.

\paragraph{Human--machine alignment.}
We next test whether the human-calibrated preference proxy recovers the human reference. We fit weights on seven annotation blocks and predict the held-out eighth, rotating through all eight blocks. Predictions cover all 1,872 comparisons, yielding Brier $0.0778$ (95\% CI: 0.0722--0.0836) and MAE $0.2115$ (0.2031--0.2200). The resulting ten-model ranking has Spearman correlation $0.915$ (0.867--0.952) with the human ranking. These internal estimates use four-vote means per judge, not the single votes used in deployment; they do not independently validate the expanded leaderboard. Page-bootstrap intervals hold these predictions fixed. Appendix~\ref{app:subjective} reports the full agreement measures, fitting protocol, and committee audit, including its pilot-to-holdout reversal.

\subsection{Subjective Leaderboard}

The subjective leaderboard measures perceived fidelity, which depends on the interaction among content, structure, and rendering errors. We therefore report it separately from the objective leaderboard rather than treating human preference as another automatic metric.

\begin{table}[htbp]
  \caption{Final subjective leaderboard from the frozen preference proxy.
  BT scores are centered; soft preference averages predicted win probability
  over scheduled appearances. Intervals use page-block bootstrap.
  The 8{,}128 additional comparisons are proxy-scored rather than
  independently human-labelled.}
  \label{tab:subjective-10000}
  \centering
  \small
  \setlength{\tabcolsep}{5pt}
  \renewcommand{\arraystretch}{1.10}

\begin{tabular}{@{} c l
      S[table-format=3.2] S[table-format=3.2] S[table-format=3.2]
      S[table-format=2.2] l @{}}
    \toprule
    {\multirow{2}{*}{\textbf{Rank}}}
      & {\multirow{2}{*}{\textbf{Model}}}
      & \multicolumn{3}{c}{\textbf{BT score (95\% CI)}}
      & {\multirow{2}{*}{\textbf{Soft pref.} $\uparrow$}}
      & {\multirow{2}{*}{\textbf{Rank CI}}} \\
    \cmidrule(lr){3-5}
    & & {\textbf{Score}} & {\textbf{Low}} & {\textbf{High}} & & \\
    \midrule
    \rowcolor{rowA} 1  & GPT 5.6 Sol            & 154.70 & 140.59 & 168.82 & 71.64\,\% & [1,\,1] \\
    \rowcolor{rowB} 2  & GPT 5.6 Terra          & 102.87 &  90.18 & 116.25 & 64.76\,\% & [2,\,3] \\
    \rowcolor{rowA} 3  & Gemini 3.1 Pro Preview & 100.86 &  89.88 & 112.74 & 64.49\,\% & [2,\,3] \\
    \rowcolor{rowB} 4  & Qwen3.7 Plus           &  56.04 &  45.63 &  66.45 & 58.12\,\% & [4,\,4] \\
    \rowcolor{rowA} 5  & Kimi K3                &   9.62 &  -2.16 &  21.71 & 51.32\,\% & [5,\,6] \\
    \rowcolor{rowB} 6  & Claude Opus 5          &  -2.47 & -14.01 &   8.96 & 49.51\,\% & [5,\,6] \\
    \rowcolor{rowA} 7  & Grok 4.5               & -55.49 & -68.68 & -43.23 & 41.77\,\% & [7,\,8] \\
    \rowcolor{rowB} 8  & Claude Sonnet 5        & -58.17 & -68.74 & -47.74 & 41.38\,\% & [7,\,8] \\
    \rowcolor{rowA} 9  & GLM 5V Turbo           & -117.28 & -129.38 & -106.41 & 33.12\,\% & [9,\,9] \\
    \rowcolor{rowB} 10 & ERNIE 5.0              & -190.68 & -203.33 & -178.59 & 23.88\,\% & [10,\,10] \\
    \bottomrule
  \end{tabular}
\end{table}

GPT-5.6-Sol ranks first in every bootstrap sample, indicating a stable advantage under the subjective protocol. The remaining systems form several overlapping tiers, especially Terra--Gemini, Kimi-K3--Opus, and Grok--Sonnet. The leaderboard therefore supports robust separation between the strongest systems and the lower-performing group, but does not justify fine-grained claims within unresolved tiers. Because the additional comparisons lack human labels, this table represents a proxy-based perceived-fidelity ranking rather than 10,000 independently human-validated comparisons.

\subsection{Comparing the Two Leaderboards}

The rankings are moderately correlated (Spearman $\rho=0.588$; Kendall $\tau=0.378$). GPT-5.6-Sol ranks first subjectively but fifth objectively, whereas Gemini ranks first objectively and third subjectively.

\begin{figure}[H]
\centering
\includegraphics[width=0.75\linewidth]{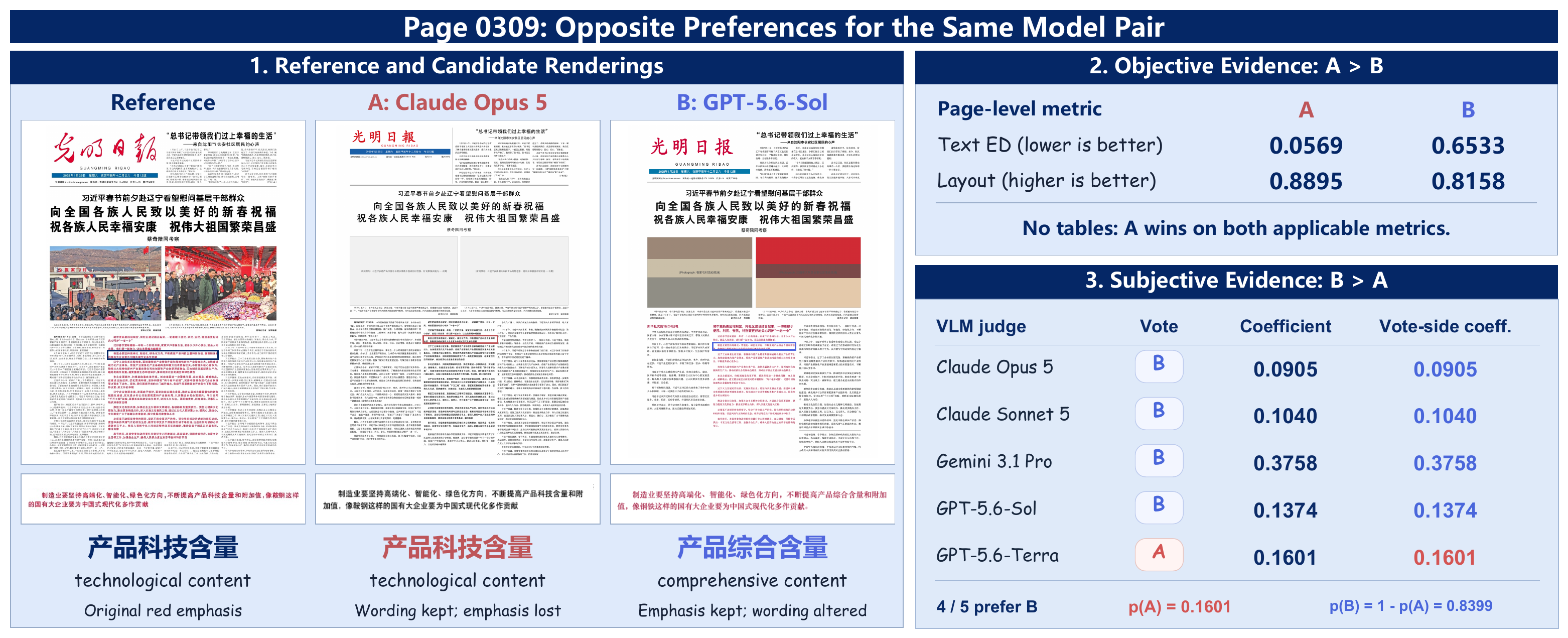}
\caption{Page-level reversal on page 0309: objective metrics favor Opus (A), whereas recorded subjective judgments favor Sol (B). Insets contrast faithful wording with preserved visual emphasis.}
\label{fig:case-level-reversal}
\end{figure}

On page 0309 (Figure~\ref{fig:case-level-reversal}), Opus wins both applicable objective metrics, but the same pair's recorded subjective proxy favors Sol (0.8399; four of five VLM votes). Sol preserves red emphasis while changing \emph{technological content} to \emph{comprehensive content}; Opus preserves the wording but loses the emphasis. This content error coexists with stronger perceived presentation, illustrating complementary evaluation signals.

\section{Conclusion}

We introduced \textbf{AmbiHTMLBench}, a 1,115-page benchmark for reconstructing heterogeneous documents as executable HTML. Its annotations and evaluation protocol treat content, structure, and rendered appearance as related but distinct requirements. The objective leaderboard shows that the best model depends on the axis being measured, while the subjective study demonstrates that perceived fidelity cannot be recovered from any single automatic component.

\clearpage
\bibliographystyle{IEEEtran}
\bibliography{paper}

\clearpage
\appendix
\section{Objective Evaluation Details}
	\label{app:objective-details}
	
	This appendix records implementation choices that are fixed for reproducibility but are not central to the paper's conceptual contribution.
	
	\subsection{Rendered block extraction and normalization}
	
	Each predicted HTML file is rendered in headless Chromium at the benchmark viewport. The evaluator traverses visible DOM content and records the semantic type, visible text, raw table HTML when available, LaTeX-like markup when exposed by the rendered DOM, and the rendered box. A predicted table is kept as one table object; its cells are not independently promoted to page-level blocks unless the table remains unmatched and is transferred to the text pool. Elements whose class attribute contains \texttt{placeholder} are excluded from matching. Invisible text, transparent text, and extremely small text are also ignored. This prevents accessibility or scene-description placeholders from contributing text that is not visibly rendered as document content.
	
	Ground-truth boxes and predicted boxes are mapped to the common page coordinate system. For each object, the outermost coverage box is used. The overlap used by matching is
	\[
	s_{\mathrm{ov}}(g,p)=
	\frac{|b_g\cap b_p|}
	{\min(|b_g|,|b_p|)}
	\]
	with zero returned for disjoint boxes. Thus, the denominator is not the union area: a small block fully covered by a larger block can obtain an overlap of one. This is useful for detecting splits and merges, but the final text and layout scores still penalize missing or excessive content.
	
	\subsection{Table assignment and TEDS approximation}
	
	For every ground-truth/predicted table pair, the evaluator computes TEDS, TEDS-S, partial text similarity, and the smaller-area overlap. The candidate score is
	\[
	0.45\,\mathrm{TEDS}
	+0.25\,\mathrm{TEDS\text{-}S}
	+0.20\,s_{\mathrm{partial}}
	+0.10\,s_{\mathrm{ov}}.
	\]
	Pairs below the table candidate threshold \(0.40\) are unavailable to the assignment. The remaining score matrix is solved with a Hungarian one-to-one assignment, followed by threshold validation. This is a global assignment over table candidates, not a greedy sort.
	
	The table parser converts HTML \texttt{tr}/\texttt{td}/\texttt{th} elements into ordered rows and cells. The implemented score is a normalized row/cell sequence edit approximation rather than a general-purpose DOM tree edit distance. TEDS retains cell strings; TEDS-S replaces them with structural cell tokens. Both are reported in \([0,1]\), with one denoting exact agreement. Cell substitutions are weighted by the normalized character-level text distance, so a one-character difference has a smaller cost than a substantially different cell. Row, insertion, deletion, and cell-structure differences are also represented. Pages without a ground-truth table are omitted from the table means.
	
	\subsection{Text normalization and grouping}

Text matching uses a conservative normalization. Whitespace and dollar delimiters are removed, while punctuation, mathematical symbols, CJK characters, Japanese, Korean, and other Unicode letters are retained. A ground-truth \(\xi\) is interpreted as a wildcard character during edit-distance computation. Placeholder elements are removed before this stage. Formula objects from both sides are converted to text objects by removing dollar delimiters, and no separate formula CDM score is computed in the current objective leaderboard.

For every ground-truth and predicted text pair, the evaluator computes \texttt{partial\_ratio} after the above normalization. A bipartite edge is created when
\[
s_{\mathrm{partial}} \ge 0.66 .
\]
Unlike earlier overlap-driven variants, this candidate graph is text-led: geometry is used later for layout scoring rather than for admitting text edges. Connected components of the bipartite graph form local candidate regions. Singleton blocks without edges are kept as residual insertions or deletions, so they still affect page text distance but do not create layout nodes.

For each connected component, the evaluator constructs several deterministic order hypotheses: visual reading order, horizontal order, reversed reading order, and a graph-guided order that places a block near its strongest partners on the other side. It chooses the pair of orders with the lowest concatenated component edit distance. For components whose total number of blocks is small enough, the evaluator additionally tries a bounded local reorder pass over the recovered groups.

Given fixed ground-truth and predicted orders, the evaluator computes the final groups with a monotone segmentation dynamic program. Each transition may consume a short ground-truth segment, a short predicted segment, or both. Two-sided transitions are supported by the bipartite edges in the component, while one-sided transitions represent insertions or deletions and receive full edit cost. The recovered groups therefore support one-to-one, one-to-many, many-to-one, and many-to-many matches while preventing unsupported distant text from being grouped solely to reduce local cost.

After all components and residual blocks are processed, the evaluator performs one page-level ordering pass over the recovered groups and computes a single normalized edit distance:
\[
D_{\mathrm{text}}=
\frac{\operatorname{Lev}(T_g,T_p)}
{\max(|T_g|,|T_p|,1)}.
\]
This max denominator makes the score symmetric with respect to which side contains more characters. A lower score is better. The same recovered groups, excluding groups with an empty side, are passed to the layout scorer.

\subsection{Layout graph similarity}
	
	The layout graph is built only from matched objects. A matched table contributes one node. A text group contributes one node whose box is the axis-aligned envelope of all GT or predicted boxes in that group. Text groups with an empty GT side or an empty prediction side are not inserted into the graph. When non-text anchors such as matched tables are available, very weak text groups whose normalized text distance exceeds \(0.75\) are also excluded from graph construction. Node similarity combines normalized center displacement with width and height agreement. For the ground-truth nodes, the evaluator computes the \(k=3\) nearest neighbors by Euclidean center distance. The same node identities are used to read the corresponding predicted relations, so the graph comparison does not require a second text matching step.
	
	For an edge \(i\rightarrow j\), direction similarity compares the normalized two-dimensional vectors, length similarity compares the relative Euclidean distances, and vector similarity compares the normalized displacement after prediction. The edge score is
	\[
	s_{\mathrm{edge}} =
	0.4s_{\mathrm{dir}}+0.3s_{\mathrm{len}}+0.3s_{\mathrm{vec}}.
	\]
	The final score is
	\[
	S_{\mathrm{layout}} =
	0.55\,\overline{s}_{\mathrm{edge}}
	+0.30\,\overline{s}_{\mathrm{node}}
	+0.15F_1^{\mathrm{coverage}}.
	\]
	This construction preserves two-dimensional information: a relation that has the correct distance but points in a different direction receives a lower score. A group with many source blocks still contributes one graph node because its envelope represents the layout region that was matched. The coverage term is computed from the matched graph nodes relative to the objects eligible for layout scoring, so pages that recover only a small subset of meaningful matched objects receive lower layout scores even when the recovered subset is geometrically accurate.
	
	\subsection{Leaderboard aggregation}
	
	The reported table means use only the 263 pages with ground-truth tables. Text error and layout similarity use all 1,115 pages. We convert text error to quality by \(1-D_{\mathrm{text}}\) and use
	\[
	S_{\mathrm{obj}} =
	\frac{0.3\,\overline{\mathrm{TEDS}}
		+\overline{(1-D_{\mathrm{text}})}
		+\overline{S_{\mathrm{layout}}}}{2.3}.
	\]
	TEDS-S is reported as a diagnostic and is not an additional term in the weighted aggregate, avoiding double-counting table structure. The final leaderboard incorporates the rerun outputs for the previously identified low-scoring pages before aggregation.

\section{Subjective Evaluation Supplement}
\label{app:subjective}
\subsection{Human study and preference targets}
\label{app:human-study}

The human study comprises eight disjoint blocks of 234 physical comparisons. Blocks 1--5 have five participants each, block 6 has four, and blocks 7--8 have two, for 33 participants in total. Each participant answers 234 main trials and six hidden repeats. Thus, the 7,920 responses comprise 7,722 main annotations and 198 repeated annotations; repeats are used to assess short-term consistency and are excluded from the calibration targets. Of the main annotations, 7,633 contain a valid A, B, or TIE preference and 89 are \texttt{CANNOT\_JUDGE}, giving 98.85\% annotation coverage. All 1,872 physical comparisons retain a human target.

For comparison $i$, $n_{i,A}$, $n_{i,T}$, and $n_{i,B}$ count valid votes after mapping the displayed candidates to their canonical identities. Their encoded mean $h_i$ measures preference for canonical A. A tie contributes half a preference point to each candidate. For example, three A votes, one tie, and one B vote give $h_i=0.7$. A value of $0.5$ can arise from unanimous ties or opposing A/B votes, so it does not by itself measure agreement. We concatenate the eight 234-entry vectors to obtain the 1,872-entry human reference; we do not average unrelated comparisons across blocks.

The reported pairwise inter-rater agreement uses 13,265 valid within-comparison rating pairs. Exact agreement is 62.30\% (95\% CI: 60.79--63.82). Fleiss-type $\kappa$ is 0.357 (0.334--0.379), and ordinal Krippendorff-type $\alpha$ is 0.459 (0.433--0.484). These quantify agreement, not error against an objective ground truth. Among the 1,872 comparisons, 39.85\% are unanimous among valid raters and 82.00\% have a unique plurality; a plurality need not exceed half the valid votes. Hidden-repeat agreement is 145/198 (73.23\%). Removing each of the 33 participants in turn leaves all ten human ranks unchanged. These checks address different sources of variation; they do not establish that individual human judgments are noise-free. Block content and participant composition are partly confounded, particularly for the two-rater blocks.

\subsection{Judge inputs, axes, and prompt}
\label{app:judge-prompt}

Each VLM receives the reference image followed by the two candidate images, with model identities omitted. The reference is the visual standard. Media neutralization removes photograph/chart interior content while retaining presence, geometry, placement, and caption relationships. Text is judged by visible presentation and substantial omissions or garbling rather than character-by-character OCR correctness. Table~\ref{tab:subjective-axes} gives the active comparison counts in the human set; multiple axes may occur in one comparison.

\begin{table}[!htbp]
\caption{Diagnostic axes in the 1,872-comparison human set. Counts denote comparisons, not distinct pages. The rarest axes support case analysis but not precise population comparisons.}
\label{tab:subjective-axes}
\centering
\small
\begin{tabular}{llrp{6.0cm}}
\hline
Activation & Axis & $n$ & Visual criteria \\
\hline
Fixed & Global layout & 1,872 & Regions, columns, spacing, grouping, reading order, proportions. \\
Fixed & Text presentation & 1,872 & Wrapping, typography, density, alignment, hierarchy, legibility. \\
Conditional & Table & 497 & Visible row/column organization and cell alignment. \\
Conditional & Formula & 61 & Fractions, scripts, radicals, matrices, and mathematical alignment. \\
Conditional & Algorithm & 12 & Block organization, indentation, and step hierarchy. \\
Conditional & Vertical text & 90 & Vertical flow, orientation, and reading organization. \\
Conditional & List hierarchy & 9 & Nesting, indentation, markers, and grouping. \\
\hline
\end{tabular}
\end{table}

The following is the shared English prompt, with line wrapping adjusted for readability:

\begin{quote}
\small
You are an independent visual reconstruction judge for AmbiHTMLBench.

You receive exactly three page images in this order: REFERENCE, CANDIDATE A, CANDIDATE B. The Reference is the sole visual standard. Determine which candidate more faithfully and completely reconstructs it as a human viewer would.

Images and charts: do not judge their internal semantic content. Judge only presence or absence, geometry, size, aspect ratio, position, alignment, and relationships to captions and neighboring regions.

Text: judge visible presentation, not character-by-character OCR correctness. Consider missing or extra text regions, large garbling, wrapping, line count, density, typography, size, weight, color, emphasis, spacing, alignment, hierarchy and legibility. Tiny character or punctuation differences that are difficult to notice should have little or no effect.

\texttt{GLOBAL\_LAYOUT\_FIDELITY} covers page-level spatial organization, regions and columns, alignment, adjacency, spacing, grouping, nesting, reading order, proportions, aspect ratios, whitespace and visual balance.

\texttt{TEXT\_VISUAL\_PRESENTATION} covers the visible presentation of text across the page.

Adaptive axes are present only when the corresponding structure exists in the Reference. Use each supplied English description and \texttt{bbox\_1000} target only to locate the Reference structure. The candidate structure may move; judge its corresponding visible organization rather than absolute coordinates.

Score every active axis: 2 = A clearly better; 1 = A slightly better; 0 = no stable perceptible difference; $-1$ = B slightly better; $-2$ = B clearly better.

Overall winner must be A, B, or TIE. It is a holistic visual judgment, not a mechanical sum of axis scores. Confidence 1--5 means expected reproducibility, not magnitude of difference.

Return strict JSON only:

\begin{verbatim}
{"winner":"A|B|TIE", "confidence":1,
 "reason":"brief English reason, <=80 words",
 "axes":[{"id":"AXIS_ID", "score":0}]}
\end{verbatim}

The axes array must exactly match the supplied active axes in membership and order. Do not mention model identities or infer which model produced a candidate.
\end{quote}

An \texttt{ACTIVE\_AXES} JSON array is appended to this prompt for each comparison. Every entry contains its axis ID, global/adaptive scope, English description, and any reference targets with their role and \texttt{bbox\_1000} coordinates. This supplies location guidance without requiring the candidate structure to occupy the same coordinates. The response validator checks the overall label, confidence range, axis membership/order, and integer scores. Swapped displays are mapped back by exchanging A/B and negating axis scores; TIE is unchanged. Confidence and free-text rationales do not enter the preference proxy.

\subsection{Preference-proxy fitting and internal validation}
\label{app:ols5}

The five judges are Claude Opus 5, Claude Sonnet 5, Gemini 3.1 Pro Preview, GPT-5.6-Sol, and GPT-5.6-Terra. Each evaluates every human comparison in two orders and two repetitions, yielding 37,440 completed outputs. Let $v_{ijor}\in\{0,0.5,1\}$ be judge $j$'s canonicalized overall vote under order $o$ and repeat $r$. The training feature is
\[
\bar x_{ij}=\frac{1}{4}\sum_{o=1}^{2}\sum_{r=1}^{2}v_{ijor},
\qquad
\widehat{\boldsymbol\beta}
=\arg\min_{\boldsymbol\beta\in\mathbb R^5}
\sum_{i=1}^{1872}(h_i-\bar{\boldsymbol x}_i^{\mathsf T}\boldsymbol\beta)^2.
\]
We use ordinary least squares, giving every physical comparison equal weight. The fit has no intercept, ridge penalty, nonnegativity constraint, or sum-to-one constraint. The resulting coefficients happen to be positive; they are not normalized. Table~\ref{tab:ols5-coefficients} records the frozen full-data coefficients. Special-axis scores do not enter this fit. In the ten-parameter representation, the other five coefficients are zero.

\begin{table}[!htbp]
\caption{Frozen preference-proxy coefficients. The deployment applies clipping to the weighted sum; the coefficients are not normalized.}
\label{tab:ols5-coefficients}
\centering
\small
\begin{tabular}{lr}
\hline
Judge & Coefficient $\beta_j$ \\
\hline
Claude Opus 5 & 0.09045204815695966 \\
Claude Sonnet 5 & 0.10401091278902799 \\
Gemini 3.1 Pro Preview & 0.37576678499354577 \\
GPT-5.6-Sol & 0.13742077090225210 \\
GPT-5.6-Terra & 0.16012498207363500 \\
\hline
\end{tabular}
\end{table}

For internal evaluation, each of the eight annotation blocks is held out in turn: 1,638 comparisons fit the coefficients and 234 receive predictions. The identifiers \texttt{PAPER-01} through \texttt{PAPER-08} denote these annotation blocks, not source research articles. The 1,872 out-of-fold predictions yield equal-case Brier/MSE $N^{-1}\sum_i(\hat p_i-h_i)^2=0.0778$ (95\% CI: 0.0722--0.0836) and MAE $0.2115$ (0.2031--0.2200), with 100\% prediction coverage. Spearman $0.915$ (0.867--0.952) compares the ten-model proxy and human rankings; it is not a case-level correlation. The full-data coefficients are used for deployment, not to compute these out-of-fold alignment estimates.

Deployment substitutes one canonicalized vote for each four-vote mean. The internal estimates therefore do not directly validate single-pass deployment on new comparisons. Also, $\sum_j\beta_j\simeq0.8678$: re-encoding all inputs as $1-x_j$ does not generally give $1-\hat p(A)$. Canonical identities are consequently fixed throughout scoring. This asymmetry is a limitation of the frozen estimator, not corrected post hoc.

\subsection{Discrete committee audit and cost ablations}
\label{app:committee-audit}

The historical committee audit uses the same repeated evidence but a different aggregation rule. A judge is resolved when at least three of its four canonicalized overall labels agree. Three judges then use two-of-three voting. Unresolved comparisons receive $p(A)=0.5$ when computing Brier and remain in its denominator. Resolved coverage is reported separately. The audit weights each comparison by its number of valid human votes divided by five, whereas the preference proxy gives each comparison equal weight; their reported losses should not be interpreted as a controlled comparison differing only in aggregation.

Eligible committees contain one judge from each vendor family and achieve at least 75\% coverage. The pilot uses 320 comparisons; the remaining 1,552 form the holdout. Selection minimizes abstention-aware Brier, breaking ties by resolved-case Brier and then coverage. Table~\ref{tab:committee-audit} reports all four eligible cross-family committees.

\begin{table}[!htbp]
\caption{Historical cross-family committee audit. Pilot $n=320$; holdout $n=1,552$. Coverage is the fraction resolved. O, S, G, L, and T denote Opus, Sonnet, Gemini, Sol, and Terra. These discrete committees are separate from the deployed preference proxy.}
\label{tab:committee-audit}
\centering
\small
\begin{tabular}{lrrr|rrr}
\hline
& \multicolumn{3}{c|}{Pilot} & \multicolumn{3}{c}{Holdout} \\
Committee & Brier & Coverage & Rank & Brier & Coverage & Rank \\
\hline
S+G+T (v1) & 0.1258 & 77.81\% & 1 & 0.1293 & 78.99\% & 4 \\
O+G+T & 0.1302 & 78.44\% & 2 & 0.1263 & 79.90\% & 2 \\
S+G+L & 0.1369 & 79.06\% & 3 & 0.1267 & 79.45\% & 3 \\
O+G+L (v2) & 0.1404 & 79.06\% & 4 & 0.1240 & 80.35\% & 1 \\
\hline
\end{tabular}
\end{table}

The pilot winner v1 falls to fourth on holdout. The paired holdout Brier difference (v1 minus v2) is $+0.0053$ (95\% CI: $-0.0001$ to $+0.0106$). Among 10,000 bootstrap resamples, v2 ranks first 76.81\% of the time. The interval includes zero, and v2 was selected after inspecting holdout; this is selection evidence rather than independent confirmation of v2. Furthermore, formula, algorithm, list, and vertical-text comparisons were concentrated in the pilot and are absent from holdout. On the full human set, descriptive macro Brier over the three special axes with at least 30 comparisons favors v1 (0.1268 versus 0.1324). Thus, the overall ordering does not establish universal superiority across structures. Gemini is present in every eligible committee, and family-specific preferences remain a potential confound.

A separate nested ablation tests all 31 nonempty judge subsets using a learned blend of overall and axis evidence. It fits nonnegative, sum-to-one weights with ridge shrinkage toward equal weights; subset selection uses only inner folds, and loss is weighted by valid human votes. With 100\% coverage on 1,872 comparisons, three judges yield Brier $0.0653$ (0.0609--0.0699); two judges yield $0.0663$ (0.0616--0.0713). The paired two-minus-three difference is $+0.0010$ ($-0.0002$ to $+0.0021$), and the one-standard-error rule selects Gemini+Sol in all eight outer folds. Five judges are worse than three by $+0.0025$ ($+0.0009$ to $+0.0041$). Intervals use 10,000 page-block resamples. These results concern an axis-aware model with different weighting and fitting, so they neither establish five judges as the minimum nor replace the frozen overall-only regression estimator. With no prespecified noninferiority margin, overlapping intervals do not prove equivalence.

\subsection{Deployment schedule and candidate versions}
\label{app:subjective-deployment}

The fixed schedule contains 10,000 comparisons across 1,113 eligible pages (seed 20260831). Each of ten models appears 2,000 times: 35 model pairs occur 222 times and ten occur 223 times. It retains the 1,872 human-study page/model pairs and adds 8,128 pairs. Pages and document structures are not sampled equally.

Each comparison receives one output per judge in a single display order, totaling 50,000 outputs. Historical directions depend on completed-result availability, never on predicted winners; reuse may therefore preserve latency-related selection effects. New requests use the primary order. The initial deployment reuses 11,380 outputs and obtains 38,620 new ones. The final table has five outputs per comparison, with no missing or duplicated entries.

The reported leaderboard incorporates the subsequent candidate synchronization. Of 58 rerun page/model outputs touched by the schedule, 54 pass the existing visible-text and render-geometry checks; four retain their earlier candidates. The replacements affect 98 comparisons: 490 judge outputs are refreshed and 49,510 retained. Pair identities, quotas, orientations, and coefficients remain fixed. All ten point ranks are unchanged, although scores change; Sol moves from 154.34 to 154.70. This checks version sensitivity, not independent performance.

Twenty-one affected comparisons come from the human-study schedule; their earlier labels do not validate the replacement candidates. The final table therefore contains 1,851 unchanged human-study comparisons, 21 updated comparisons without new human labels, and 8,128 unlabeled additions. The original calibration set is unchanged. Neither its overlap with deployment nor the updated table supplies external human validation.

The initial deployment's final 22 failures comprise 14 request-size failures and eight timeouts. Eighteen repaired calls use high-quality JPEG reference copies; one further reduces the long edge from 6,320 to 6,000 pixels (5.1\%), preserving aspect ratio, to meet the visual-patch limit. Original references remain unchanged. These repairs are separate from candidate replacements.

\subsection{Ranking and uncertainty}
\label{app:subjective-bootstrap}

For comparison $i$ between models $a_i$ and $b_i$, the proxy outcome contributes $\hat p_i(A)$ preference points to $a_i$ and $1-\hat p_i(A)$ to $b_i$. The soft-preference column is the mean of these points over a model's 2,000 scheduled appearances. It is not accuracy against human labels.

Bradley--Terry fitting estimates centered log-abilities $\theta_m$ using
\[
q_i=\sigma(\theta_{a_i}-\theta_{b_i}),\qquad
\max_{\boldsymbol\theta}\;
\sum_i\big[\hat p_i(A)\log q_i+(1-\hat p_i(A))\log(1-q_i)\big]
-\frac{10^{-3}}{2}\sum_m\theta_m^2.
\]
The small $\ell_2$ term stabilizes the ranking fit; it is not a penalty on the OLS coefficients. Centering sets the mean ability to zero. Reported scores are $400\theta_m/\log(10)$, an Elo-like relative scale. A negative score indicates below-average estimated ability within this comparison pool, not failed reconstruction.

For the deployment leaderboard, 2,000 resamples draw 1,113 page blocks with replacement (seed 20260831). Each selected page carries all of its scheduled comparisons. The Bradley--Terry model is refitted for every resample; 2.5th and 97.5th percentiles form the score and rank intervals. These are marginal intervals and do not constitute simultaneous tests of every model pair. ``First in every bootstrap sample'' describes the observed resampling frequency, not certainty under all sources of error.

Proxy alignment uses 5,000 page-block resamples stratified by annotation block (seed 20260830), with fixed out-of-fold predictions; the committee audit uses 10,000 similarly stratified resamples. Within each stratum, comparisons from the same page remain together. Bootstrap fixes proxy coefficients and predictions, omitting renewed judge sampling, regression fitting/selection, candidate generation/cleaning, and systematic bias. Alignment is internal to the human study; the expanded leaderboard provides no full-scale human Accuracy, Recall, or F1.

\clearpage
\section{Case Study}
\label{app:case study}

The reason for performing poorly on AmbiHTMLBench could vary. Some typical cases are listed below.

\begin{figure}[h]
\centering
\includegraphics[width=.65\linewidth]{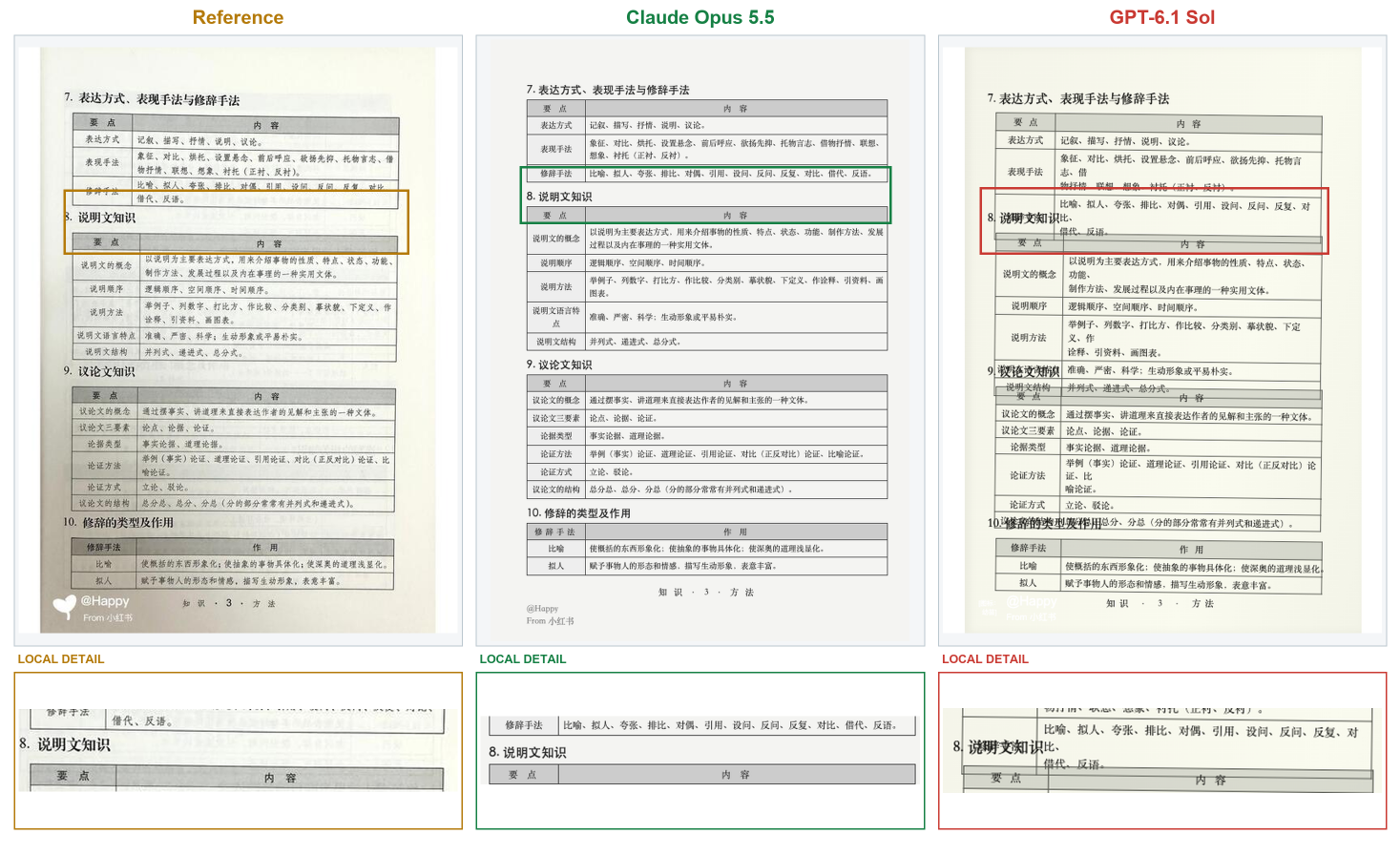}
\caption{\textbf{Text occlusion.} Claude Opus 5.5 keeps the section heading between the two tables. In GPT-6.1 Sol, the heading overlaps the preceding table, disrupting the separation between sections. This makes the document difficult to read visually in terms of layout.}
\label{fig:case-header-occlusion}
\end{figure}
\begin{figure}[h]
\centering
\includegraphics[width=.65\linewidth]{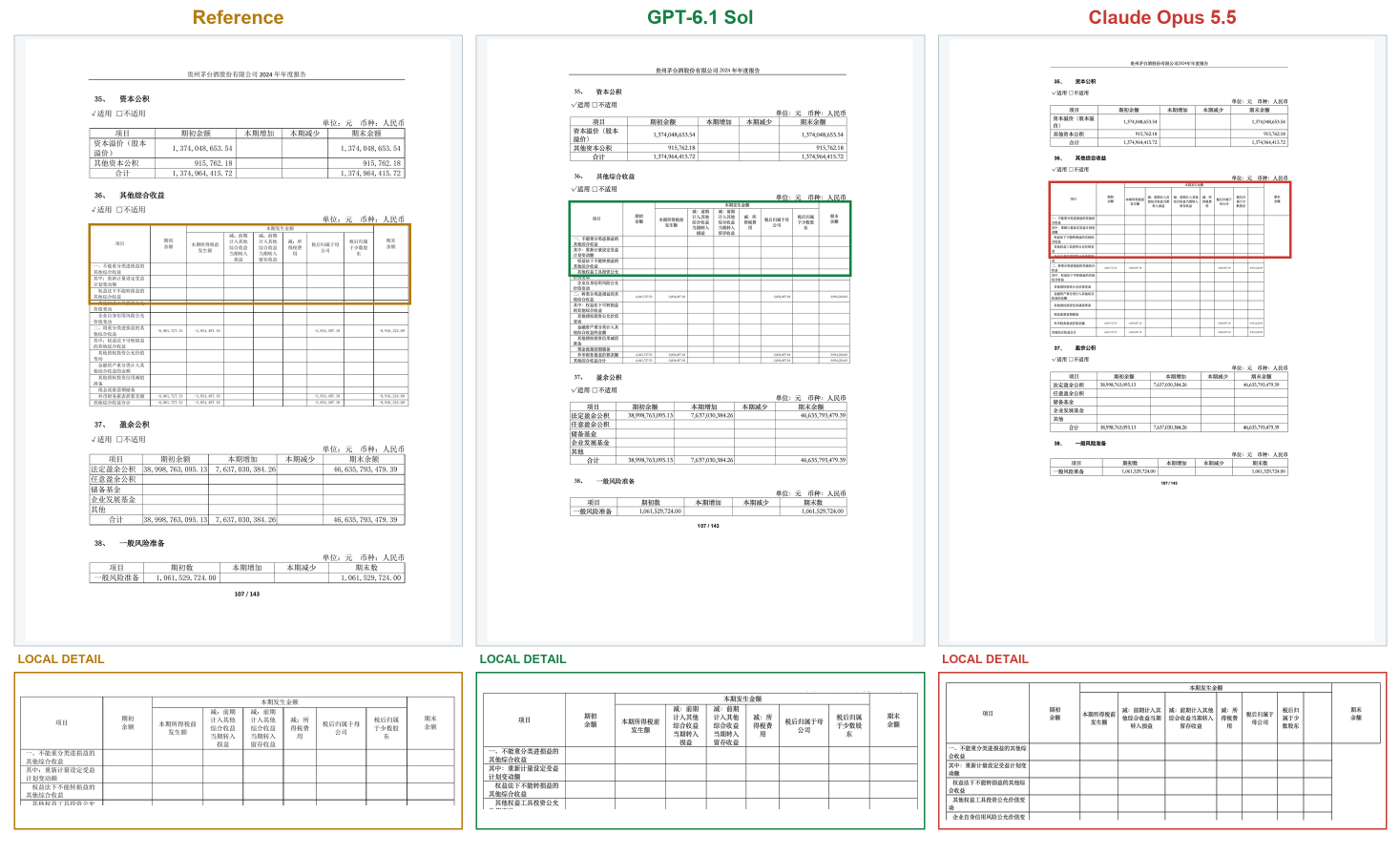}
\caption{\textbf{Table structure degradation.} GPT-6.1 Sol aligns the final column header with the table body. In Claude Opus 5.5, the header extends beyond the right edge of the body, leaving the column misaligned, and the header of the second-to-last column has not been properly recognized.}
\label{fig:case-table-boundaries}
\end{figure}

\begin{figure}[h]
\centering
\includegraphics[width=.8\linewidth]{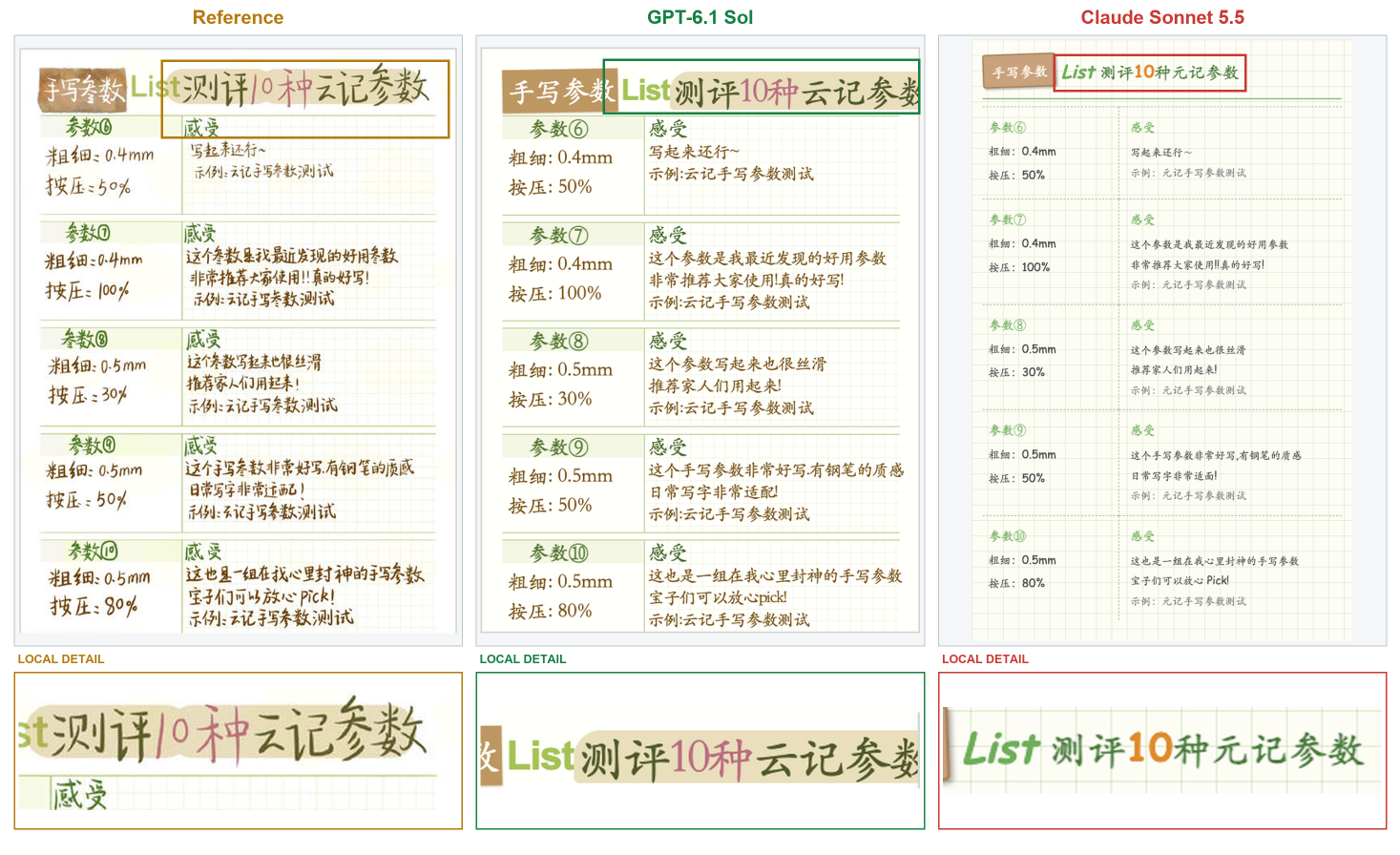}
\caption{\textbf{Text reproduction error.} Claude Sonnet 5.5 substitutes a different character; The distinctive color and doodle style of social media notes have also been eliminated. In this case, GPT-6.1 Sol towers over.}
\label{fig:case-vertical-text}
\end{figure}

\begin{figure}[h]
\centering
\includegraphics[width=.8\linewidth]{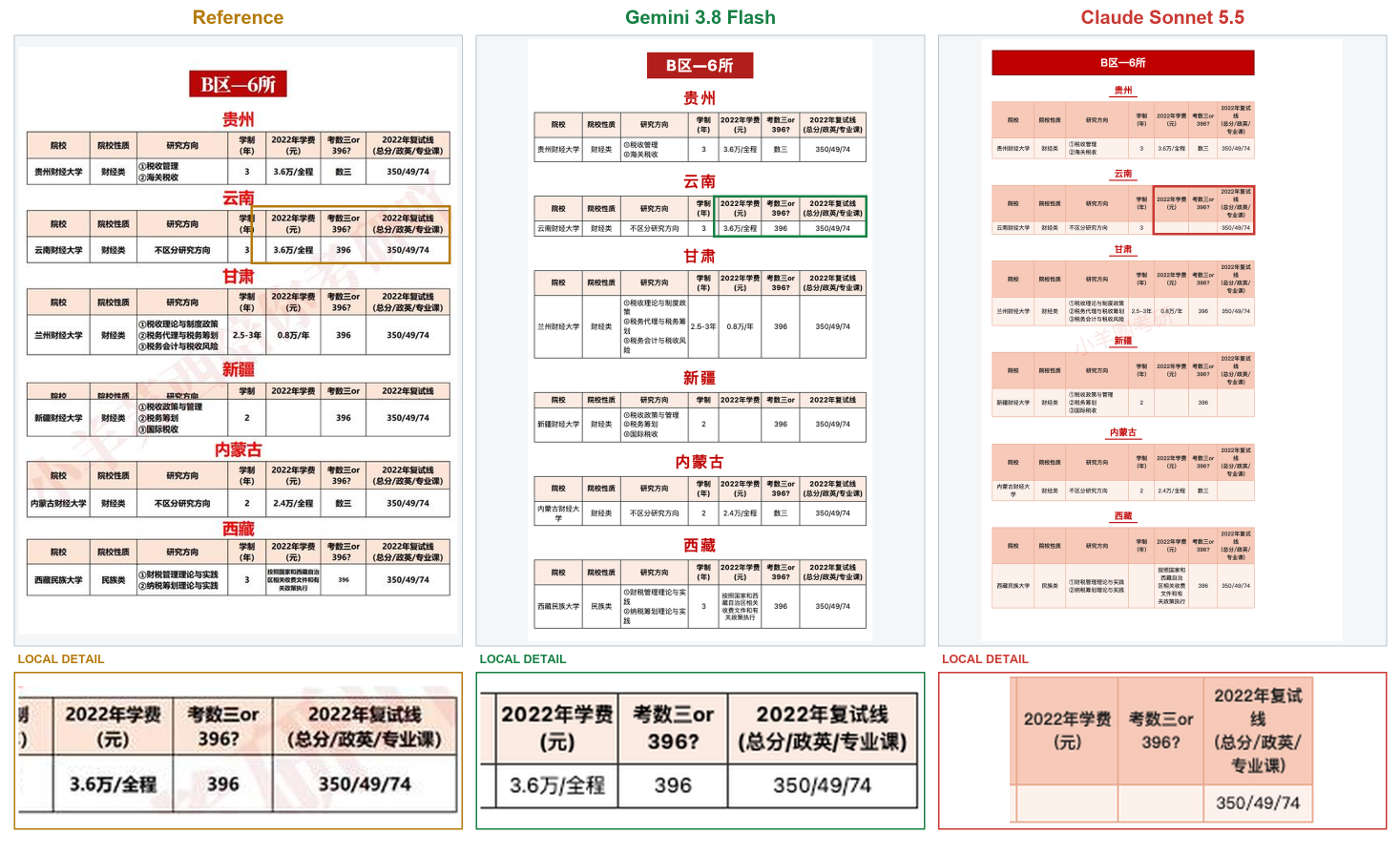}
\caption{\textbf{Information loss.} Gemini 3.8 Flash retains the tuition and examination information in the highlighted row. Claude Sonnet 5.5 leaves both corresponding cells empty.}
\label{fig:case-paragraph-omission}
\end{figure}

\begin{figure}[h]
\centering
\includegraphics[width=.8\linewidth]{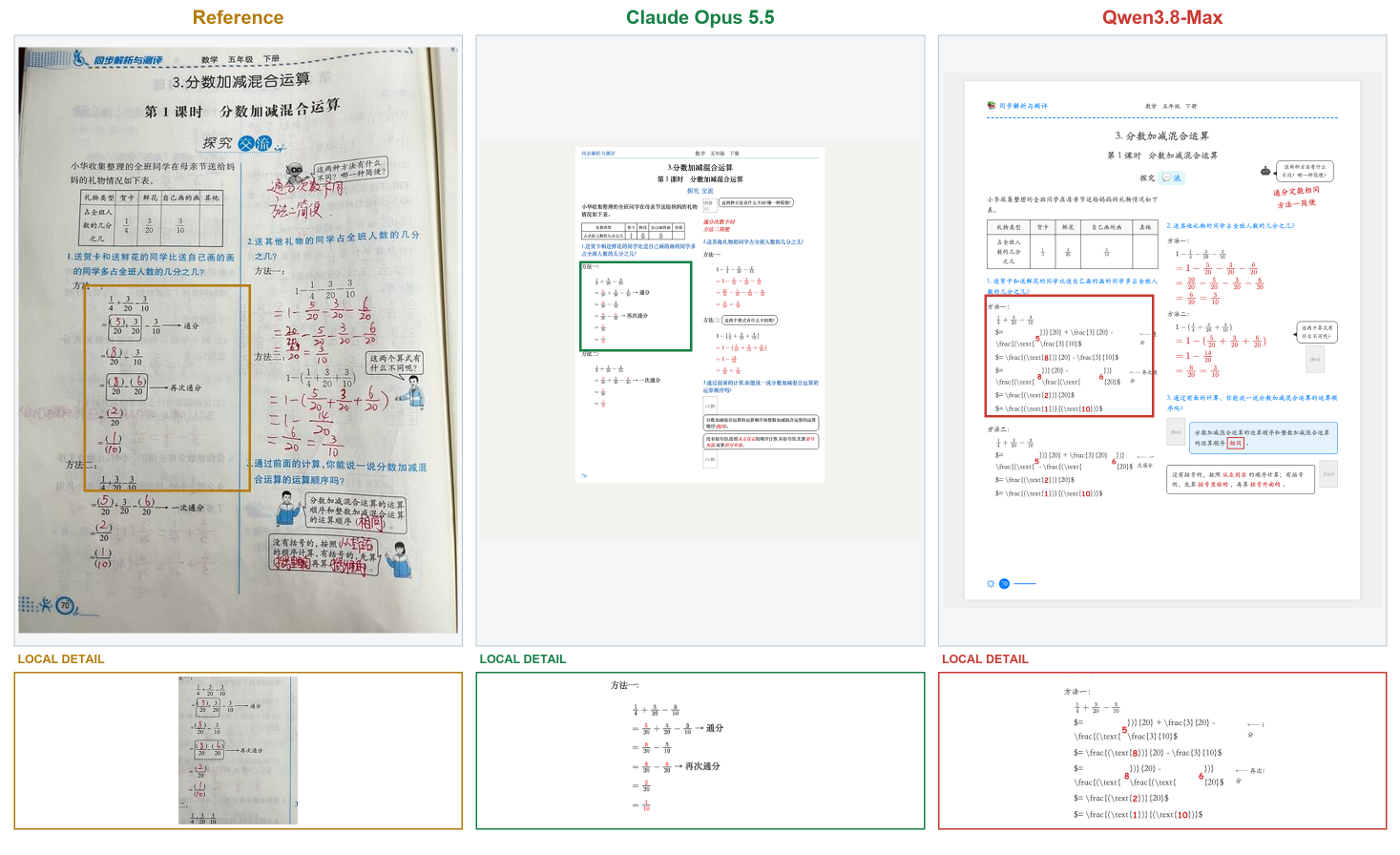}
\caption{\textbf{Formula rendering error.} Claude Opus 5.5 renders the highlighted calculation as two-dimensional fractions. Qwen3.8-Max displays raw formula commands in several lines, obscuring the mathematical structure.}
\label{fig:case-formula-substitution}
\end{figure}

\begin{figure}[h]
\centering
\includegraphics[width=.8\linewidth]{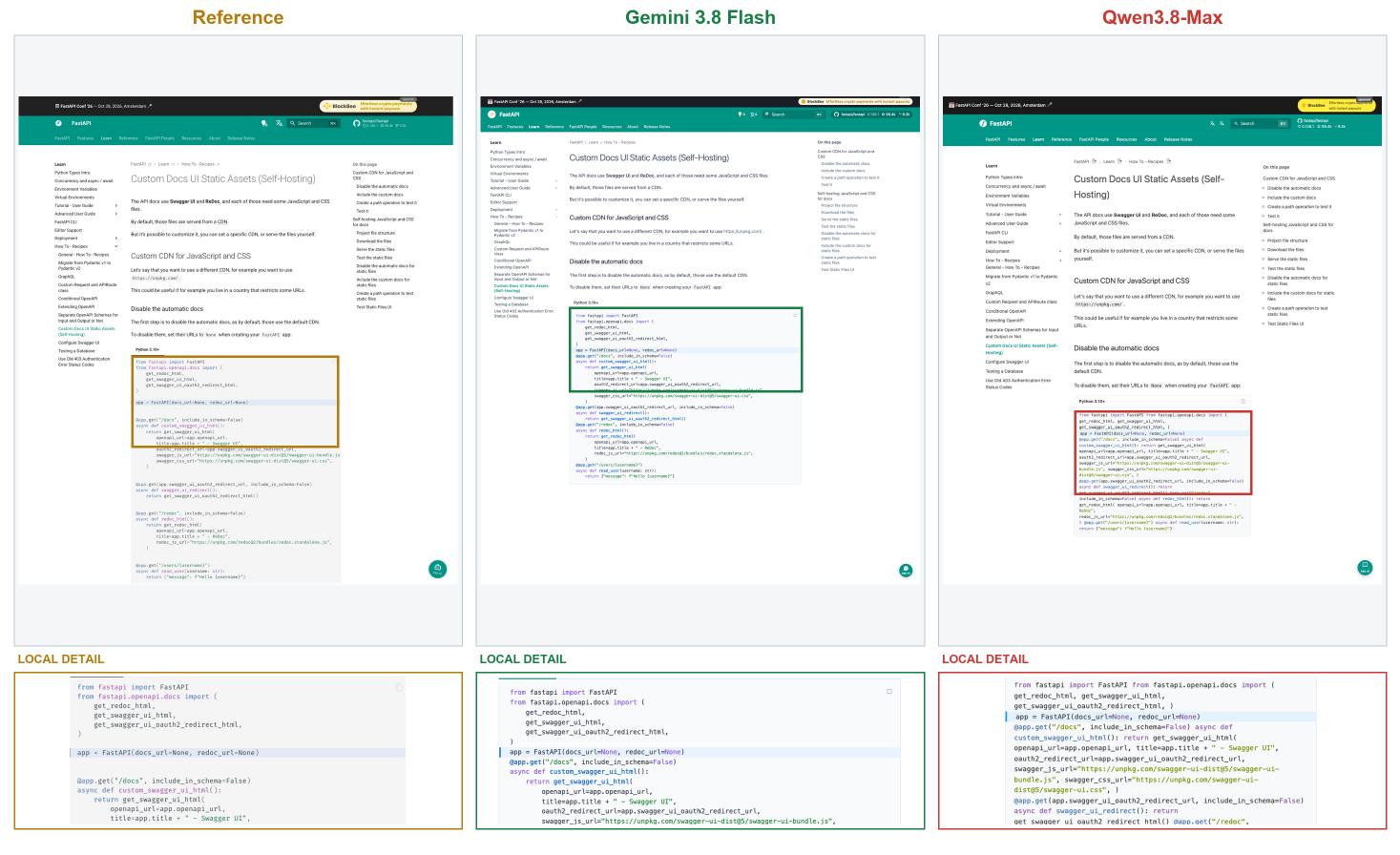}
\caption{\textbf{Code formatting error.} Gemini 3.8 Flash retains the line breaks and indentation of the reference code block. Qwen3.8-Max merges separate statements and loses indentation, making the code structure harder to read.}
\label{fig:case-address-substitution}
\end{figure}
\end{document}